 \pdfoutput=1
\documentclass[11pt,twoside]{article}
\usepackage{newtitle}
\usepackage{fullpage}
\usepackage{epsfig}
\usepackage{url}
\usepackage{hyperref}
\usepackage{graphics}

\hypersetup{%
	pdftitle      = {Simulated Car Racing Championship: Competition Software Manual},
	pdfsubject    = {},
	pdfkeywords   = {},
	pdfauthor     = {Daniele Loiacono, Luigi Cardamone, Pier Luca Lanzi},
	pdfcreator    = {\LaTeX\ with package \flqq hyperref\frqq},
	colorlinks = true,
	linkcolor = red,
	anchorcolor = blue,
	citecolor = blue,
	filecolor = blue,
	urlcolor = blue,
	hypertexnames=false,
	pagecolor = blue,
}

\title{Simulated Car Racing Championship\\
Competition Software Manual}
\author{Daniele Loiacono, Luigi Cardamone, Pier Luca Lanzi\\
Politecnico di Milano, Italy\\
\texttt{\{loiacono,cardamone,lanzi\}@elet.polimi.it}}
\author{Daniele Loiacono, \texttt{daniele.loiacono@polimi.it}\\
Luigi Cardamone, \texttt{luigi.cardamone@polimi.it}\\
Pier Luca Lanzi, \texttt{pierluca.lanzi@polimi.it}\\
~\\	
Politecnico di Milano, Dipartimento di Elettronica, Informazione e Bioingegneria, Italy}

\date{April 2013}

\newcommand{\TORCSPackage}{\mbox{\tt scr-}}
\newcommand{\TORCSBot}{\mbox{\tt scr\_server}}
\newcommand{\TORCSCurrent}{\mbox{TORCS 1.3.4}}
\newcommand{\TORCSClient}{\mbox{\tt scr-client-cpp}}
\newcommand{\TORCSClientJava}{\mbox{\tt scr-client-java}}
\begin{document}

\maketitle

\title{Simulated Car Racing Championship\\
Competition Software Manual}

\author{Daniele Loiacono, Politecnico di Milano, Italy, \texttt{loiacono@elet.polimi.it}\\
Luigi Cardamone, Politecnico di Milano, Italy, \texttt{cardamone@elet.polimi.it}\\
Pier Luca Lanzi, Politecnico di Milano, Italy, \texttt{lanzi@elet.polimi.it}}

\begin{abstract}
This manual describes the competition software for the Simulated Car Racing Championship, an international competition
held at major conferences in the field of Evolutionary Computation and in the field of Computational Intelligence and Games.
It provides an overview of the architecture, the instructions to install the software and to run the simple drivers provided in the package, the description of the sensors and the actuators.



\end{abstract}

\section{Introduction}
This manual describes the competition software for the Simulated Car Racing Championship, an international competition
held at major conferences in the field of Evolutionary Computation and in the field of Computational Intelligence and Games.

The goal of the competition is to design a controller for a racing car that will compete on a set of unknown tracks first alone (against the clock) and then against other drivers.
The controllers perceive the racing environment through a number of sensors that describe the relevant features of the car surroundings (e.g., the track limits, the position of near-by obstacles), of the car state (the fuel level, the engine RPMs, the current gear, etc.), and the current game state (lap time, number of lap, etc.). The controller can perform the typical driving actions (clutch, changing gear, accelerate, break, steering the wheel, etc.).
A description of the championship, including the rules and regulations, can be found at \url{http://cig.sourceforge.net/}

The championship platform is built on top of The Open Racing Car Simulator (\href{http://www.torcs.org}{TORCS}) 
a state-of-the-art open source car racing simulator which provides a full 3D visualization,
a sophisticated physics engine, and accurate car dynamics taking into account traction, aerodynamics, fuel consumption, etc.

In the remainder of this manual, we provide an overview of the architecture, the instructions to install the software and to run the simple drivers provided in the package, the description of the sensors and the actuators.

\section{The Architecture of the Competition Software}
\label{sec:architecture}
The Open Racing Car Simulator (TORCS) comes as a stand-alone application in which the bots are compiled as separate modules that are loaded into main memory when a race takes place.
This structure has three major drawbacks. First, races are not in real-time since bots execution is blocking: if a bot takes a long time to decide what to do, it will block all the others.
Second, since there is no separation between the bots and the simulation engine, the bots have full access to all the data structures defining the track and the current status of the race.
As a consequence, each bot can use different information for its driving strategy. Furthermore,  bots can analyze the complete state of the race (e.g., the track structure, the opponents position, speed, etc.) to plan their actions. Accordingly, a fair comparison among methods of computational intelligence is difficult since different methods might access different information.
Last but not least, TORCS restricts the choice of the programming language to C/C++ since the bots must be compiled as loadable module of the main TORCS application which is written in C++.

The competition software extends the original TORCS architecture in three respects. 
First, it structures TORCS as a client-server applications: the bots are run as external processes connected to the race server through \href{http://en.wikipedia.org/wiki/User_Datagram_Protocol}{UDP} connections.
Second, it adds real-time: every game tic (roughly corresponding to 20ms of simulated time), the server sends the current sensory inputs to each bot and then it waits for 10ms (of real time) to receive an action from the bot. If no action arrives, the simulation continues and the last performed action is used.
Finally, the competition software creates a physical separation between the driver code and the race server building an abstraction layer, a \emph{sensors and actuators model}, which (i) gives complete freedom of choice regarding the programming language used for bots and (ii) restricts the access only to the information defined by the designer.

\begin{figure}[t]
\centerline{\includegraphics{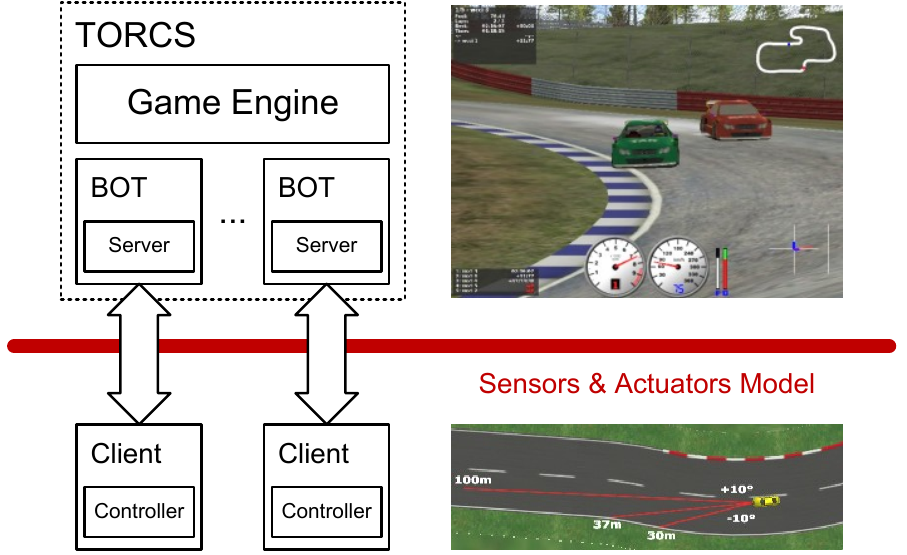}}
\caption{The architecture of the competition software.}
\label{fig:architecture}
\end{figure}

The architecture of the competition software is shown in Figure~\ref{fig:architecture}. The game engine is the same as the original TORCS, the main modification in a new \emph{server-bot}, called \TORCSBot, which manages the connection between the game and a client bot using UDP.
A race involves one server-bot for each client; each server-bot listens on a separate port of the race server.
At the beginning, each client-bot identifies itself to a corresponding \emph{server-bot} establishing a connection. Then, as the race starts, each server-bot sends the current sensory information to its client and awaits for an action until 10ms (of real time) have passed. Every game tic, corresponding to 20ms of simulated time, the server updates the state of the race which is sent back to the clients.
A client can request a race restart by sending a special action to the server.

\section{Installing the Competition Server}
\label{sec:install}
To provide a very accessible interface to TORCS we developed two modules to run TORCS in a client/server architecture. The server has been developed by providing a specific bot driver called {\tt \TORCSBot} that, instead of having its own intelligence, sends the game state to a client module and waits for a reply, i.e., an action to be performed by the controller.
So to begin the competition, we first need to install TORCS and the competition server-package
  provided in this bundle.

\subsection{Linux Version}
Download the all-in-one \TORCSCurrent\ source package from SourceForge (\url{http://sourceforge.net/projects/torcs/}) 
or directly from 
\href{http://prdownloads.sourceforge.net/torcs/torcs-1.3.4.tar.bz2?download}{here}.
To compile the server you will need:
\begin{itemize}
\item Hardware accelerated OpenGL (usually provided by your Linux distribution)
\item GLUT 3.7 or FreeGlut (better than GLUT for full screen support)
\item PLIB 1.8.5 version
\item OpenAL
\item libpng and zlib (usually provided by your Linux distribution)
\item FreeALUT
\end{itemize}

Unpack the package \texttt{torcs-1.3.4.tar.bz2}, 
with ``{\tt tar xfvj torcs-1.3.4.tar.bz2}" which will create the directory \texttt{torcs-1.3.4}. 

Download the package \TORCSPackage{\tt linux-patch.tgz} containing the patch for the TORCS sources from the CIG project page at \url{http://sourceforge.net/projects/cig/} or as a direct download from \href{https://sourceforge.net/projects/cig/files/SCR%20Championship/Server%20Linux/}{here}.
Unpack the package \TORCSPackage{\tt linux-patch.tgz} in your base TORCS directory (where you unpacked \texttt{torcs-1.3.4.tar.bz2}).
This will create a new directory called {\tt \TORCSPackage patch}.
Enter the {\tt \TORCSPackage patch} directory and run the script {\tt do\_patch.sh} with ``{\tt sh do\_patch.sh}" 
(run {\tt do\_unpatch.sh} to revert the modifications). 
Move to the parent directory (where you unpacked \texttt{torcs-1.3.4.tar.bz2}) and run
\begin{verbatim}
      $ ./configure        
      $ make
      $ make install
      $ make datainstall
\end{verbatim}
At this point you should be able to check whether the competition software has been properly installed by executing the command ``{\tt torcs}"; then, from the main window select
\[ 
 \mbox{Race $\rightarrow$ Quick Race $\rightarrow$ Configure Race $\rightarrow$ Accept}.
\]
If everything has been installed correctly, you should find ten instances of the 
  \TORCSBot\ bot in the list of ``Not Selected Player'' on the righthand side.

\medskip\noindent
Further information about the installation process are available  \href{http://torcs.sourceforge.net/index.php?name=Sections&op=viewarticle&artid=3#linux-src-all}{here}.
Additional information are available at \url{http://www.berniw.org/} (TORCS $\rightarrow$ Installation  -- from left bar).

\subsection{Windows Version}

It is possible to compile TORCS on Windows from sources but it can be rather challenging. 
Therefore, we provide the binary distribution of the competition software for Windows. 
In this case, to install the competition software first download the \TORCSCurrent\ Windows installer from \url{https://sourceforge.net/projects/torcs/files/torcs-win32-bin/} 
and install it.
Then, download the file \TORCSPackage{\tt win-patch.zip} from the CIG project page at \url{http://sourceforge.net/projects/cig/} or as a direct download from \href{https://sourceforge.net/projects/cig/files/SCR%20Championship/Server%20Windows/}{here}. Unzip the package in the TORCS main directory. 
During the unpacking, you will be asked to overwrite some existing files, answer yes for all the files.
At this point you should be able to check whether the competition software has been properly installed by launching {\tt wtorcs.exe} from the installation directory or from the start menu; then, from the TORCS main window select,
\[
 \mbox{Race $\rightarrow$ Quick Race $\rightarrow$ Configure Race $\rightarrow$ Accept}.
\]
If everything has been installed correctly, you should find ten instances of the 
  \TORCSBot\ bot in the list of ``Not Selected Player'' on the righthand side.

\subsection{Mac OsX}
We do not provide support for Mac OsX since it is not supported by the TORCS developers.

\section{The C++ Client}
\label{sec:client}
The C++ client for the competition is a stand-alone console applications that can be compiled from the sources.
The package can be downloaded from the CIG project page at \url{http://sourceforge.net/projects/cig/} or as a direct download from \href{https://sourceforge.net/projects/cig/files/SCR%20Championship/Client%20C%2B%2B/}{here}.

\subsection{Compiling for Linux}
Unpack the client package \TORCSClient{\tt .tgz} creating the directory \TORCSClient.
Then, open a terminal in the directory where you unpacked the client and type 
  \verb|make| to compile it.
The compilation process should end without any error or warning and you should now have 
  an executable named \verb|client| in you directory.
To launch the client, type,
\begin{verbatim}
      $ ./client  host:<ip>  port:<p> id:<client-id> maxEpisodes:<me> \ 
		  maxSteps:<ms> track:<trackname> stage:<s>
\end{verbatim}
where \verb|<ip>| is the IP address of the machine where the TORCS competition server is running 
(the default is \verb|localhost|);
  \verb|<p>| is the port on which the server-bot is listening, typical values are between \verb|3001| and \verb|3010|
(the default is \verb|3001|);
  \verb|<client-id>| is your bot ID 
(the default is  \texttt{SCR});
\verb|<me>| is the maximum number of learning episodes to perform 
(the default value is  \texttt{1});
\verb|<ms>| is the maximum number of control steps in each episode.
(the default value is  \texttt{0}, i.e., unlimited number of steps);
\verb|<trackname>| is the name\footnote{This trackname is \emph{not} necessarily the name of the track in TORCS. 
It is a conventional name that can be used by the bot to store its own information specific for the different tracks.}
  of the track where the bot will race (default value is \texttt{unknown});
\verb|<s>| is an integer that represents the current stage of the competition the bot is involved in: 
0 is \emph{Warm-up}, 1 is \emph{Qualifying}, 2 is \emph{Race}, and 3 is \emph{Unknown} (default is \emph{Unknown}).
All the parameters are optional (if not specified, the default values are used instead).

\subsection{Compiling for Windows}
Unpack the client package \TORCSClient{\tt .tgz} creating the directory \TORCSClient.
The package is provided with a DevC++ project file (\url{http://www.bloodshed.net/devcpp.html}) 
  but any C++ development tool/IDE may be used.
To compile the client on Windows, uncomment the first two lines of \verb|client.cpp| 
  following the instructions provided in the same file.
The package also contains the system library \verb|WS2_32.lib| that is required for using 
  the \emph{WinSock} functions.
The client should compile without any error or warning, producing  
  the \verb|client.exe| executable.
To launch the client open an MS-DOS console and type:
\begin{verbatim}
client.exe host:<ip>  port:<p> id:<client-id> maxEpisodes:<me> \ 
		  maxSteps:<ms> track:<trackname> stage:<s>
\end{verbatim}
where \verb|<ip>| is the IP address of the machine where the TORCS competition server is running 
(the default is \verb|localhost|);
  \verb|<p>| is the port on which the server-bot is listening, typical values are between \verb|3001| and \verb|3010|
(the default is \verb|3001|);
  \verb|<client-id>| is your bot ID 
(the default is  \texttt{SCR});
\verb|<me>| is the maximum number of learning episodes to perform 
(the default value is  \texttt{1});
\verb|<ms>| is the maximum number of control steps in each episode.
(the default value is  \texttt{0}, i.e., unlimited number of steps);
\verb|<trackname>| is the name\footnote{This trackname is \emph{not} necessarily the name of the track in TORCS. 
It is a conventional name that can be used by the bot to store its own information specific for the different tracks.}
  of the track where the bot will race (default value is \texttt{unknown});
\verb|<s>| is an integer that represents the current stage of the competition the bot is involved in: 
0 is \emph{Warm-up}, 1 is \emph{Qualifying}, 2 is \emph{Race}, and 3 is \emph{Unknown} (default is \emph{Unknown}).
All the parameters are optional (if not specified, the default values are used instead).

\subsection{Customizing Your Own Driver}
To write your own driver, the \verb|BaseDriver| class provided in the client sources must be extended 
and these methods must be implemented:
\begin{itemize}
\item \verb|void init(float *angles)|,
  the method is called before the beginning of the race and can be used to define a custom configuration of the \emph{track} 
  sensors (see Table~\ref{tab:sensors-II}): 
the desired angles (w.r.t. the car axis) of all the 19 range finder sensors must be set in the parameter \texttt{angles}.
\item \verb|string drive(string sensors)|,
  where \texttt{sensors} represents the current state of the game as perceived by 
  your driver; the method returns a string representing the actions taken (see Section~\ref{sec:api} for details regarding sensors and actuators);
\item \verb|void onShutdown()|, 
  the method called at the end of the race, before the driver module is unloaded;
\item \verb|void onRestart()|,
   the method called when the race is restarted upon the driver request
   (this function should be used to free allocated memory, close open files, saving to disk, etc.).
\end{itemize}
In addition, the class attributes \verb|stage| and \verb|trackName| contain respectively the current stage of the race (\emph{warm-up}, \emph{qualifying}, \emph{race} or \emph{unknown})
and the name of the current track (both the \emph{stage} and the \emph{track name} must be specified using the corresponding command line option of the client). 
This information can be used to save useful information about the current track as well as to adopt different strategies in the different stages of the competition.
As an example, the file \verb|SimpleDriver.cpp| implements a very simple driver and 
it is used by default to build the client executable.
Therefore, to build a client executable to run your own driver:
\begin{itemize}
\item \textbf{on Windows}, uncomment the first two lines of \verb|client.cpp| and set the
 \verb|__DRIVER_CLASS__| and \verb|__DRIVER_INCLUDE__| definitions to the name of the implemented driver class and to 
  the header file of the same driver class.
\item \textbf{on Linux}, in the \verb|Makefile| set \verb|DRIVER_CLASS| and \verb|DRIVER_INCLUDE| to the name of the implemented driver class and to 
  the header file of the same driver class.
\end{itemize}

\section{The Java Client}
\label{sec:java_client}
The Java client works similarly to the C++ version. 
It is a stand-alone console application that can be compiled from the sources. 
The package can be downloaded from the CIG project page at \url{http://sourceforge.net/projects/cig/} or as a direct download from \href{https://sourceforge.net/projects/cig/files/SCR%20Championship/Client%20Java/}{here}.
%
%
\subsection{Running the Java Client}
First, unpack the package \TORCSClientJava{\tt .tgz} to create the directory \TORCSClientJava\ containing the source code.
To compile the client, go to the directory \texttt{src} and type,
\begin{verbatim}
      $ javac -d ../classes scr/*.java
\end{verbatim}
%
To launch the Java client with a simple controller, go to the directory \texttt{classes} and type,
%
\begin{verbatim}
      $ java scr.Client scr.SimpleDriver host:<ip> port:<p> id:<client-id> \
        maxEpisodes:<me> maxSteps:<ms> verbose:<v> track:<trackname> stage:<s>
\end{verbatim}
%
where \verb|scr.SimpleDriver| is the implementation of a controller provided with the software as an example 
  (it can be replaced with a custom implementation);
where \verb|<ip>| is the IP address of the machine where the TORCS competition server is running 
(the default is \verb|localhost|);
  \verb|<p>| is the port on which the server-bot is listening, typical values are between \verb|3001| and \verb|3010|
(the default is \verb|3001|);
  \verb|<client-id>| is your bot ID 
(the default is  \texttt{SCR});
\verb|<me>| is the maximum number of learning episodes to perform 
(the default value is  \texttt{1});
\verb|<ms>| is the maximum number of control steps in each episode.
(the default value is  \texttt{0}, i.e., unlimited number of steps);
\verb|<v>| controls the verbosity level, it can be either \texttt{on} or \texttt{off}
(the default value is \texttt{off});
\verb|<trackname>| is the name\footnote{This trackname is \emph{not} necessarily the name of the track in TORCS. 
It is a conventional name that can be used by the bot to store its own information specific for the different tracks.}
  of the track where the bot will race (default value is \texttt{unknown});
\verb|<s>| is an integer that represents the current stage of the competition the bot is involved in: 
0 is \emph{Warm-up}, 1 is \emph{Qualifying}, 2 is \emph{Race}, and 3 is \emph{Unknown} (default is \emph{Unknown}).
All the parameters are optional (if not specified, the default values are used instead).

\subsection{Customizing Your Own Driver}
The Java client is organized similarly to the C++ client.
To write your own driver, the \texttt{Controller} interface must be implemented by providing the following methods:
\begin{itemize}
\item \verb|public float[] initAngles()|.
  the method is called before the beginning of the race and can be used to define a custom configuration of the \emph{track} 
  sensors (see Table~\ref{tab:sensors-II}): 
the method returns a vector of the 19 desired angles (w.r.t. the car axis) for each one of the 19 range finders.
\item \verb|public Action control(SensorModel sensors)|,
  where \texttt{sensors} represents the current state of the game as perceived by 
  your driver; the method returns the action taken (see Section~\ref{sec:api});
\item \verb|public void shutdown()|, 
  the method called at the end of the race, before the driver module is unloaded;
\item \verb|public void reset()|,
   the method called when the race is restarted upon the driver request
   (this function should be used to free allocated memory, close open files, saving to disk, etc.).
\end{itemize}
In addition, the class attributes \verb|stage| and \verb|trackName| contains respectively the current stage of the race (\emph{warm-up}, \emph{qualifying}, \emph{race} or \emph{unknown})
and the name of the current track (both the \emph{stage} and the \emph{track name} must be specified using the corresponding command line option of the client).
This information can be used to save useful information about the current track as well as to adopt different strategies in the different stages of the competition.
As an example, the file \verb|SimpleDriver.java| implements a very simple driver.

\section{Sensors and Actuators}
\label{sec:api}
The competition software creates a physical separation between the game engine and the drivers. 
Thus, to develop a bot it is not required any knowledge about the TORCS engine or the internal data structure.
The drivers perceptions and the available actions are defined by a sensors and actuators layer defined by the competition designer. For this competition, the drivers inputs consists of some data about the car status (the current gear, the fuel level, etc.) the race status (the current lap, the distance raced, etc.) and the car surroundings (the track borders, the obstacles, etc.). The actions allow the typical driving actions.

\subsection{Sensors}
The bot perceives the racing environment through a number of sensor readings which provide information both 
  about the surrounding game environment (e.g., the tracks, the opponents, the speed, etc.) and the current 
  state of the race (e.g., the current lap time and the position in the race, etc.).
Table~\ref{tab:sensors-I} and Table~\ref{tab:sensors-II} report the complete list of sensors available along with a description.
Please notice that the readings provided by \verb|opponents| sensors (Table~\ref{tab:sensors-I}) 
  do not take into account the edges of the track, i.e., 
  distances between cars are computed ``as the crow flies'' even if the paths 
  cross the edges of the track.
\begin{table}[ht]
\begin{center}
\begin{tabular}{|c|c|l|}
\hline
\textbf{Name} & \textbf{Range (unit)} & \textbf{Description} \\ \hline
angle & [-$\pi$,+$\pi$] (rad) &
\parbox{0.6\textwidth}{\vspace{0.05in}
Angle between the car direction and the direction of the track axis.
\vspace{0.05in}} \\ \hline
curLapTime & [0,$+\infty$) (s) &
\parbox{0.6\textwidth}{\vspace{0.05in}
Time elapsed during current lap.
\vspace{0.05in}} \\ \hline
damage & [0,$+\infty$) (point) &
\parbox{0.6\textwidth}{\vspace{0.05in}
Current damage of the car (the higher is the value the higher is the damage).
\vspace{0.05in}} \\ \hline
distFromStart & [0,$+\infty$) (m) & 
\parbox{0.6\textwidth}{\vspace{0.05in}
Distance of the car from the start line along the track line.
}\\ \hline
distRaced & [0,$+\infty$) (m) &
\parbox{0.6\textwidth}{\vspace{0.05in}
Distance covered by the car from the beginning of the race
\vspace{0.05in}} \\ \hline
focus & [0,200] (m) &
\parbox{0.6\textwidth}{\vspace{0.05in}
Vector of 5 range finder sensors: 
  each sensor returns the distance between the track edge and the car within a range of 200 meters.
When \texttt{noisy} option is enabled (see Section~\ref{sec:server}) sensors are affected by i.i.d. normal noises with a standard deviation equal to the 1\% of sensors range.
The sensors sample, with a resolution of one degree, a five degree space along a specific direction provided by the client 
  (the direction is defined with the \emph{focus} command and must be in the range [-90,+90] degrees w.r.t. 
  the car axis).
Focus sensors are not always available: they can be used only once per second of simulated time.
When the car is outside of the track (i.e., pos is less than -1 or greater than 1), the focus direction is outside the allowed range
  ([-90,+90] degrees) or the sensors has been already used once in the last second, the returned values are not reliable (typically -1 is returned).
\vspace{0.05in}} \\ \hline
fuel & [0,$+\infty$) (l) &
\parbox{0.6\textwidth}{\vspace{0.05in}
Current fuel level.
\vspace{0.05in}} \\ \hline
gear & \{-1,0,1,$\cdots$ 6\} &
\parbox{0.6\textwidth}{\vspace{0.05in}
Current gear: -1 is reverse, 0 is neutral and the gear from 1 to 6.
\vspace{0.05in}} \\ \hline
lastLapTime & [0,$+\infty$) (s) &
\parbox{0.6\textwidth}{\vspace{0.05in}
Time to complete the last lap
\vspace{0.05in}} \\ \hline
opponents & [0,200] (m) &
\parbox{0.6\textwidth}{\vspace{0.05in}
Vector of 36 opponent sensors:
each sensor covers a span of 10 degrees within a range of 200 meters and returns the distance of the 
  closest opponent in the covered area.
When \texttt{noisy} option is enabled (see Section~\ref{sec:server}), sensors are affected by i.i.d. normal noises with a standard deviation equal to the 2\% of sensors range.
The 36 sensors cover all the space around the car, spanning clockwise from -180 degrees up to +180 degrees with respect 
  to the car axis.
\vspace{0.05in}} \\ \hline
racePos & \{1,2,$\cdots$,$N$\} &
\parbox{0.6\textwidth}{\vspace{0.05in}
Position in the race with respect to other cars.
\vspace{0.05in}} \\ \hline  
rpm & [0,$+\infty$) (rpm) &
\parbox{0.6\textwidth}{\vspace{0.05in}
Number of rotation per minute of the car engine.
\vspace{0.05in}} \\ \hline
speedX & ($-\infty$,$+\infty$) (km/h) &
\parbox{0.6\textwidth}{\vspace{0.05in}
Speed of the car along the longitudinal axis of the car.
\vspace{0.05in}} \\ \hline
speedY & ($-\infty$,$+\infty$) (km/h) &
\parbox{0.6\textwidth}{\vspace{0.05in}
Speed of the car along the transverse axis of the car.
\vspace{0.05in}} \\ \hline
speedZ & ($-\infty$,$+\infty$) (km/h) &
\parbox{0.6\textwidth}{\vspace{0.05in}
Speed of the car along the Z axis of the car.
\vspace{0.05in}} \\ \hline
\end{tabular}
\caption{Description of  the available sensors (part I). Ranges are reported with their unit of measure (where defined).}
\label{tab:sensors-I}
\end{center}
\end{table}

\begin{table}[ht]
\begin{center}
\begin{tabular}{|c|c|l|}
\hline
track & [0,200] (m) &
\parbox{0.6\textwidth}{\vspace{0.05in}
Vector of 19 range finder sensors: 
  each sensors returns the distance between the track edge and the car within a range of 200 meters.
When \texttt{noisy} option is enabled (see Section~\ref{sec:server}), sensors are affected by i.i.d. normal noises with a standard deviation equal to the 10\% of sensors range.
By default, the sensors sample the space in front of the car every 10 degrees, 
  spanning clockwise from -90 degrees up to +90 degrees with respect to the car axis.
However, the configuration of the range finder sensors (i.e., the angle w.r.t. to the car axis) can be set by the client once during initialization, i.e., before the beginning of each race.
When the car is outside of the track (i.e., pos is less than -1 or greater than 1), the returned values are 
  not reliable (typically -1 is returned).
\vspace{0.05in}} \\ \hline
trackPos & ($-\infty$,$+\infty$) & 
\parbox{0.6\textwidth}{\vspace{0.05in}
Distance between the car and the track axis.
The value is normalized w.r.t to the track width: it is 0 when car is on the axis, -1 when the car is 
on the right edge of the track and +1 when it is on the left edge of the car. 
Values greater than 1 or smaller than -1 mean that the car is outside of the track.
\vspace{0.05in}} \\ \hline
wheelSpinVel & [0,$+\infty$] (rad/s) &
\parbox{0.6\textwidth}{\vspace{0.05in}
Vector of 4 sensors representing the rotation speed of wheels.
\vspace{0.05in}} \\ \hline
z & [$-\infty$,$+\infty$] (m) &
\parbox{0.6\textwidth}{\vspace{0.05in}
Distance of the car mass center from the surface of the track along the Z axis.
\vspace{0.05in}} \\ \hline
\end{tabular}
\caption{Description of the available sensors (part II). Ranges are reported with their unit of measure (where defined).}
\label{tab:sensors-II}
\end{center}
\end{table}

\subsection{Actuators}
The bot controls the car in the game through a rather typical set of actuators, 
  i.e., the steering wheel, the gas pedal, the brake pedal, and the gearbox.
In addition, a \emph{meta-action} is available to request a race restart to the server.
Table~\ref{tab:effectors} details the actions available and their representation.
\begin{table}[ht]
\begin{center}
\begin{tabular}{|c|c|l|}
\hline
\textbf{Name} & \textbf{Range} & \textbf{Description} \\ \hline
accel & [0,1] &
\parbox{0.6\textwidth}{\vspace{0.05in}
Virtual gas pedal (0 means no gas, 1 full gas).
\vspace{0.05in}} \\ \hline
brake & [0,1] &
\parbox{0.6\textwidth}{\vspace{0.05in}
Virtual brake pedal (0 means no brake, 1 full brake).
\vspace{0.05in}} \\ \hline
clutch & [0,1] &
\parbox{0.6\textwidth}{\vspace{0.05in}
Virtual clutch pedal (0 means no clutch, 1 full clutch).
\vspace{0.05in}} \\ \hline
gear & -1,0,1,$\cdots$,6 &
\parbox{0.6\textwidth}{\vspace{0.05in}
Gear value.
\vspace{0.05in}} \\ \hline
steering & [-1,1] &
\parbox{0.6\textwidth}{\vspace{0.05in}
Steering value: -1 and +1 means respectively full right and left, that corresponds to an angle of 0.366519 rad.
\vspace{0.05in}} \\ \hline
focus & [-90,90] &
\parbox{0.6\textwidth}{\vspace{0.05in}
Focus direction (see the \emph{focus} sensors in Table~\ref{tab:sensors-I}) in degrees.
\vspace{0.05in}} \\ \hline
meta & 0,1 &
\parbox{0.6\textwidth}{\vspace{0.05in}
This is meta-control command: 0 do nothing, 1 ask competition server to restart the race.
\vspace{0.05in}} \\ \hline
\end{tabular}
\caption{Description of the available effectors.}
\label{tab:effectors}
\end{center}
\end{table}

\section{Running the Competition Server}
\label{sec:server}
Once you have installed TORCS and the server-bot provided (either Windows or Linux version), 
  you can start to develop your own bot extending one of the provided client modules.
When you want to run your own bot you have to launch TORCS and start a race, then
  you have to launch the client extended with your own programmed bot and 
  finally your driver bot will start to run in the race.
In TORCS there are several race modes available, however the client-server modules supports only
  two modes:
\begin{itemize}
\item the \emph{Practice} mode that allows a single bot at once to race
\item the \emph{Quick Race} modes that allows multiple bots to race against
\end{itemize}
However, before starting a race with TORCS, you need to configure the following things:
\begin{itemize}
\item you have to select the track on which you want to run the race
\item you need to add a \verb|scr_server x| bot to race participants and eventually other bots you want as opponents
\item you have to define how many laps or how many kilometers that race will last
\item you might want to select the desired display mode
\end{itemize}
In TORCS, all the above options are stored in a set of XML configuration files (one for each race mode).
Under Linux configuration files are created after the game is launched for the first time and are located in 
\verb|$HOME/.torcs/config/raceman/|, where \verb|$HOME| is your home directory.
Under Windows instead the configuration files are located in the \verb|\config\raceman\| directory located under 
  the directory where you installed TORCS.

\subsection{Configuring TORCS Race Via GUI}
The easiest way to configure the race options is using the TORCS GUI.
Each race mode can be fully configured selecting from the main menu of TORCS:
\[
 \mbox{Race $\rightarrow$ Quick Race [or Practice] $\rightarrow$ Configure Race}.
\]
Once you change the configurations of a particular race mode, all the changes are stored automatically by TORCS 
  the corresponding configuration file.
\paragraph{Selecting track.}
In the first screen you can select any of the track available in the games and then click on \emph{Accept} 
  to move to the next screen.
\paragraph{Selecting bots.}
The second screen allows the selections of bot that will participate to the race.
Notice that in the \emph{Practice} mode only one bot is allowed, therefore in order to add a bot you have first to 
  deselect the currently selected one (if any).
First of all you have to make sure that one competition bot, {\tt \TORCSBot\ x}, is in the 
  list of selected drivers (on the left of the screen).
Then, in the \emph{Quick Race} mode only, you can add other drivers to the race from the list on the right 
(representing all the bot drivers provided with the game).
When adding bots pay attention to the car model they use: there are several types of car in TORCS with different 
  features and you might want to be sure that only drivers with the same cars will race against.
Notice that \TORCSBot\ uses a {\tt car1-trb1} and the others bot using the same car are:
\begin{itemize}
\item {\tt tita 3}
\item {\tt berniw 3}
\item {\tt olethros 3}
\item {\tt lliaw 3}
\item {\tt inferno 3}
\item {\tt bt 3}
\end{itemize}
When you have selected all the drivers that will be in the race, you can click on \emph{Accept} and move
  to the next screen
\paragraph{Setting race length and display mode.}
In the final configuration screen you can set the race length either as the distance to cover (in km) or as 
  the number of laps to complete.
Finally you can choose between two display modes option: \emph{normal} or \emph{results only}.
The \emph{normal} mode allows you to see the race either from the point of view of one bot driver or as an 
  external spectator.
In this display mode, the time speed can be accelerated up to four times the normal speed, that is you can see 
  1 minute of race in 15s.
In the \emph{results only} mode instead you will not see the race but only the lap times (in \emph{Practice} mode)
  or the final result of the race (in \emph{Quick Race} mode).
However this mode allow you to run simulation much faster: time speed can be accelerated up to 20 times (or even more), 
  that is one minute of race can be simulated within 3 seconds.
%
%
\subsection{Configuring TORCS through Configuration Files}
All the race settings described above can be configured also editing directly a configuration file.
In TORCS each race type as its own XML configuration file. 
The settings of \emph{Practice} are stored in \verb|practice.xml| while the settings of \emph{Quick Race}
  are in \verb|quickrace.xml|.

\paragraph{Selecting track.}
To select the track, find the ``Tracks'' section inside the XML file, that will contain the following
  section:
\begin{verbatim}
<section name="1">
  <attstr name="name" val="TRACK-NAME"/>
  <attstr name="category" val="TRACK-CAT"/>
</section>
\end{verbatim} 
where you should 
(i) replace \verb|TRACK-ROAD| with the category of desired 
  track (i.e., \verb|road|,\verb|oval| or \verb|dirt|);
(ii) replace \verb|TRACK-NAME| with the name of desired 
  track (e.g., \verb|aalborg|).
For a complete list of the installed tracks in TORCS, you can see the list of all the directories organized under three 
  main directories, \verb|tracks/road/|,\verb|tracks/oval/| and \verb|tracks/dirt/|, where TORCS is installed.
Under Windows you find them in your main torcs directory, under Linux the tracks directories could be found in 
  \verb|/usr/local/share/games/torcs/| or in different places depending on your distribution.

\paragraph{Selecting bots.}
To select bots you should modify the ``Drivers'' section inside the XML file.
In particular in this section you should be able to find a list of the following elements:
\begin{verbatim}
<section name="N">
  <attnum name="idx" val="IDX"/>
  <attstr name="module" val="NAME"/>
</section>
\end{verbatim}
where \verb|N| means you are editing the $N$th bots that will be in the race.
The \verb|IDX| is the index of the instance of the bot you want to add: for some bots provided with the game there are 
  several instances (e.g., \verb|bt| bot has several instances: \verb|bt 1|, \verb|bt 2|, $\cdots$); 
  when a bot has only one instance \verb|IDX| should be set to 1).
The \verb|NAME| should be replaced with the name of bot you want to add without the index of the instance (e.g., 
  to add the \verb|bt 7| bot, you should use as \verb|NAME| simply \verb|bt| and \verb|7| as \verb|IDX|).
A list of available drivers can be found in the \verb|drivers/| directory located in the same place where 
  you have the \verb|tracks| directory introduced before.

\paragraph{Setting race length and display mode.}
To change race length and display mode you have to modify the ``Quick Race'' or ``Practice'' section 
  (depending on which race type you want to setup).
In particular you should change the following lines:
\begin{verbatim}
...
<attnum name="distance" unit="km" val="DIST"/>
...
<attnum name="laps" val="LAPS"/>
...
<attstr name="display mode" val="MODE"/>
...
\end{verbatim}
where \verb|DIST| should be either the desired race length in km or $0$ if the number of laps is used as 
  race length.
Accordingly, \verb|LAPS| should be either the desired number of laps or $0$ if the distance is used as race length.
Finally \verb|MODE| is either \verb|normal| or \verb|results only|.


\subsection{Start to Race!}
Once you configured properly TORCS you are ready to run your own bot.
From the main menu of TORCS select:
\[
 \mbox{Race $\rightarrow$ Quick Race [or Practice] $\rightarrow$ New Race}.
\]
You should see that TORCS screen should stop reporting the line 
\[
\verb|Initializing Driver scr_server 1...|
\]
The OS terminal should report \verb|Waiting for request on port 3001|.
This means that the server-bot \TORCSBot\ is waiting for your client to start the race.
After the race is started, it can be interrupted from the user by pressing \verb|ESC| and then by 
  selecting \verb|Abort Race| from the menu. 
The end of the race is notified to the client either if it has been interrupted by a user or 
  if the distance/lap limit of the race has been reached. 
Please notice that it the \verb|Quit Game| option is chosen in the game menu, instead of the \verb|Abort Race| 
  option, the end of the race will not be notified correctly to the clients preventing them from
  performing a clean shutdown.

\subsection{Running TORCS in text-mode}
 It is possible to run TORCS without graphics, i.e. without any GUI to launch the race.
 This run mode could be useful when you plan to run an experiment (or a series of experiments) in
   a batch mode and you do not need to use the GUI to setup the experiment.
Using the ``-r'' command line option it is possible to specify the race configuration file an to run TORCS in text-mode, 
as follows:
 \begin{verbatim}
  C:\> wtorcs.exe -r race_config.xml (on Windows)
  $ torcs -r race_config.xml (on linux)
 \end{verbatim}
TORCS will run automatically the race defined by the \verb|race_config.xml| file, that can be configured either using the GUI 
   or directly editing it (as explained in the previous section).

\subsection{Disabling Fuel, Damage and Laptime Limit}
\label{cmdlineopts}

To performs very long experiments in TORCS it is necessary to disable some features that can stop or alter the simulation. Fuel consumption and damage should be disabled for two reasons: first, they increase the noise in the evaluation process because two individuals with a different amount of fuel or damage have different performance; second if the fuel is low or the damage too high the car is removed from the race.
The laptime limit removes a car from a race if it takes to much to complete a lap. This situation can happen if with a particular configuration of parameters the car performances are very poor.

To disable these features it is possible to run the patched version of TORCS with these command line arguments:
\begin{verbatim}
      C:\> wtorcs.exe -nofuel -nodamage -nolaptime (on Windows)
      $ torcs -nofuel -nodamage -nolaptime (on Linux)
\end{verbatim}
Of course each of this arguments can be used alone or in combination with the others.

\subsection{Time Constraints}
In the development of your driver, please keep in mind that the race is in real-time. 
Accordingly, the server has a timeout on the client answers:
your driver should perform an action (i.e., return an action string) by \textbf{10ms} in order to keep in sync with the server.
If your bot is slower, you would probably loose the sync with the server and so it is up to you 
  to find out how to avoid that this will happen.
It is also possible to specify a custom timeout through the following command line option:
\begin{verbatim}
      C:\> wtorcs.exe -t <timeout> (on Windows)
      $ torcs -t <timeout> (on Linux)
\end{verbatim}
where \verb|timeout| is the desired timeout (measured as nanoseconds).

\subsection{Noisy sensors}
By default, the range finders in the sensor model are \emph{not} noisy.
However, during the competition noisy range finders will be used (according to the specification in Table~\ref{tab:sensors-I} and Table~\ref{tab:sensors-II}).
To enable noisy range finders, it is possible to use the following command line option:
\begin{verbatim}
      C:\> wtorcs.exe -noisy (on Windows)
      $ torcs -noisy (on Linux)
\end{verbatim}

\clearpage
\section{Further Information and Support}
\label{sec:further}
Further information about the championship and the competition software is available at \url{http://cig.sourceforge.net/}. 

\medskip\noindent
To report bugs, problems, or just for help, send an email to \href{mailto:scr@geccocompetitions.com}{scr@geccocompetitions.com}.

\medskip\noindent
Additional information is also available from the following websites:
\begin{itemize}
\item \url{http://www.torcs.org}, The Open Racing Car Simulator main website
\item \url{http://www.berniw.org/}, Bernhard Wymann's page with a lot of information about TORCS
\end{itemize}

\end{document}